\pdfoutput=1

\documentclass[11pt]{article}

\usepackage[preprint]{acl}

\usepackage{times}
\usepackage{latexsym}

\usepackage[T1]{fontenc}

\usepackage[utf8]{inputenc}

\usepackage{microtype}

\usepackage{inconsolata}

\usepackage{graphicx}
\usepackage{amsmath}
\usepackage{changepage}   
\usepackage[shortlabels]{enumitem}
\usepackage{lscape}
\usepackage{enumitem}
\usepackage{float}
\usepackage{amssymb}
\usepackage{hyperref}
\usepackage{cleveref}
\usepackage{booktabs}
\usepackage{adjustbox}
\usepackage{subcaption}
\usepackage{multirow}
\usepackage{orcidlink}
\usepackage{xurl}

\usepackage[dvipsnames]{xcolor}
\definecolor{Mycolor1}{HTML}{117a65}
 
\title{Evaluating the Capabilities of LLMs for Persuasive Dialogue}

\author{
  \textbf{Jordan Robinson\textsuperscript{1,2\orcidlink{0009-0001-6828-3524}}},
  \textbf{Angus R. Williams\textsuperscript{2\orcidlink{0009-0004-3136-3768}}}
  \textbf{Katie Atkinson\textsuperscript{1,2\orcidlink{0000-0002-5683-4106}}},
  \textbf{Anthony G. Cohn\textsuperscript{2,3\orcidlink{0000-0002-7652-8907}}}
\\
\\
  \textsuperscript{1}University of Liverpool,
  \textsuperscript{2}The Alan Turing Institute,
  \textsuperscript{3}University of Leeds
}

\begin{document}
\maketitle

\begin{abstract}
Large language models (LLMs) can generate apparently highly persuasive text, but does sounding persuasive mean arguing well? We introduce \textsc{Persuasio}, a multi-agent dialogue platform grounded in a formal argumentation-based theory of persuasion dialogues that adjudicates logical winners during free-text debates. Using this system, we generated 192 debates on a UK political topic between humans and LLMs, and evaluated 22 interlocutors through both automated adjudication and 9,702 crowdsourced pairwise judgements across 1{,}386 annotation instances. We observed a consistent decoupling between subjective and formal persuasiveness: LLMs dominated the subjective ranking yet performed substantially worse under argumentation-theoretic adjudication, where humans remained competitive. Multi-agent and retrieval-augmented variants further widened this divergence. These findings reveal a systematic gap between rhetorical fluency and formal argumentative strength in LLM-based persuasive dialogues.
\end{abstract}

\section{Introduction}

Persuasion has been studied for millennia, from Aristotle's \textit{Rhetoric}~\cite{2010rhetoric} to modern computational models of argumentation and persuasion~\cite{hunter2016computational}. Persuasive communication aims to shape or change the beliefs of others \cite{stiff2016persuasive}. Recent work shows that LLMs can generate arguments that humans find compelling across political, health, and social domains \cite{Artificial-Intelligence-Can-Persuade-Humans-on-Political-Issues, The-potential-of-generative-AI-for-personalized-persuasion-at-scale, repec:nat:nathum:v:9:y:2025:i:8:d:10.1038_s41562-025-02194-6, doi:10.1126/science.adq1814}. However, it remains unclear whether perceived dialogical persuasiveness corresponds to logical persuasiveness.

Existing evaluations typically assess either rhetorical appeal through human judgements \cite{Tan_2016, habernal-gurevych-2016-argument} or structural properties of arguments through automated analysis \cite{stab-gurevych-2014-identifying, wachsmuth-etal-2017-computational}, but rarely both jointly. As a result, we lack direct evidence on whether model fluency and confident framing align with inferential rigour and logical consistency. If these dimensions diverge, systems optimised for perceived persuasiveness may produce rhetorically polished yet argumentatively fragile discourse.

We address this issue by introducing \textsc{Persuasio}, a modular multi-agent dialogue platform implementing the formal persuasion dialogue system of \citet{10.1017/S0269888906000865}. The platform enforces turn-taking, commitment tracking, and formal termination conditions, enabling automated adjudication of logical winners. Using \textsc{Persuasio}, we generated 192 UK political debate transcripts spanning human--human, human--model, and model--model configurations involving 22 interlocutors (13 humans, 9 LLM variants). 

We evaluate persuasiveness along two axes: \emph{logical} persuasiveness, derived from formal dialogue adjudication, and \emph{subjective} persuasiveness, derived from 9,702 pairwise judgements across 1{,}386 samples. We model both using Bradley--Terry (BT) models \cite{bradley1952} to enable direct comparison. We find that all 9 LLM configurations occupy the top positions in subjective persuasiveness rankings, yet drop substantially w.r.t. logical persuasiveness, revealing a systematic gap between the two. This has implications for using persuasive AI in deliberative domains such as policy, law, and education, where argumentative integrity is essential. In summary, our contributions are:
\begin{enumerate}[noitemsep,wide]
    \item We introduce \textsc{Persuasio}, an open-source platform,\footnote{Source code: \url{https://github.com/RobinsonJI/Evaluating-the-Capabilities-of-LLMs-for-Persuasive-Dialogue}} formally underpinned by the work of \citet{10.1017/S0269888906000865}, allowing users to engage in persuasive dialogues with LLM agents.
    \item We release a corpus of 192 political debate transcripts between 22 distinct human and LLM-based variants, along with the 9{,}702 crowdsourced annotations and the corresponding demographic data of annotators.
    \item We evaluate the capabilities of LLMs for persuasive dialogues, providing the first direct comparison of logical and subjective persuasiveness, using rankings inferred from BT models over the same set of 22 interlocutors.
\end{enumerate}

\section{Related Work and Background}

\paragraph{LLMs for Persuasion.}
LLMs can generate personalised messages tailored to individual preferences \cite{Artificial-Intelligence-Can-Persuade-Humans-on-Political-Issues, The-potential-of-generative-AI-for-personalized-persuasion-at-scale, Breum_Egdal_Gram_Mortensen_Møller_Aiello_2024, 10.1108/ITP-10-2021-0764, Comparing-the-persuasiveness-of-role-playing-large-language-models-and-human-experts-on-polarized-U.S.-political-issues, repec:nat:nathum:v:9:y:2025:i:8:d:10.1038_s41562-025-02194-6, doi:10.1126/science.adq1814}, exploiting cognitive biases that affect human decision-making \cite{Breum_Egdal_Gram_Mortensen_Møller_Aiello_2024, 10.1108/ITP-10-2021-0764, doi:10.1126/science.adq1814, carrascofarre2024largelanguagemodelspersuasive, griffin2023susceptibilityinfluencelargelanguage}. Multi-turn interactions can be more effective than one-way messaging \cite{Breum_Egdal_Gram_Mortensen_Møller_Aiello_2024, 10.1108/ITP-10-2021-0764, doi:10.1126/science.adq1814, Information-Delivered-by-a-Chatbot-Has-a-Positive-Impact-on-COVID-19-Vaccines-Attitudes-and-Intentions, LI2026113353}, while maintaining consistency across prolonged dialogues \cite{Breum_Egdal_Gram_Mortensen_Møller_Aiello_2024, hackenburg2024evidencelogscalinglaw}. 
Closest to our work, \citet{dewynter2025linecomprehensionpersuasionllms} generate human--model debate transcripts under a formal dialogue model to assess the ability of LLMs to maintain debates across many dialogue turns, and then they utilise both human annotations and model completions to evaluate how this capability relates to their comprehension of dialogical structures and pragmatic context. Our work is distinct from this: we measure the persuasiveness logically, using the formalism of \citet{10.1017/S0269888906000865}, and subjectively, using crowdsourced annotations of debate transcripts.

\paragraph{Evaluating Persuasiveness.}
Debate platforms and online forums provide rich data for studying persuasive dynamics \cite{Tan_2016, habernal-gurevych-2016-argument, hidey-etal-2017-analyzing}. Computational models have used argumentation-based features \cite{stab-gurevych-2014-identifying} to rank persuasiveness in Reddit posts \cite{wei-etal-2016-post} and student essays \cite{persing-ng-2015-modeling, carlile-etal-2018-give}. A key question is whether persuasiveness derives from argument quality or rhetorical style. Some work links argument cogency and reasonableness to perceived persuasiveness \cite{wachsmuth-etal-2017-computational}, while others highlight the influence of style and reader ideology \cite{el-baff-etal-2020-analyzing}.

\paragraph{Argumentation-Based Persuasion Dialogues.}
Our system builds on formal dialogue-theoretic frameworks. Persuasion dialogues, among six dialogue types \cite{walton1995commitment}, involve resolving opinion conflicts through regulated argumentative exchange. We adopt the framework of \citet{10.1017/S0269888906000865}, which defines a communication language $\mathcal{L}_c$ of six locutions (\textit{claim}, \textit{why}, \textit{concede}, \textit{retract}, \textit{since}, \textit{question}), turn-taking protocols, and commitment stores to ensure consistency. Dialogue progression and termination are formalised via transformations over these commitments and the interlocutors' initial claims. Full formal details appear in Appendix~\ref{app:sec:persuasion-dialogues}, and implementation specifics in Section~\ref{sec:multi-agent-system-design}.

\section{Multi-Agent System and Dialogue Game Design}\label{sec:multi-agent-system-design}

\begin{figure}[h!]
    \centering
    \includegraphics[width=0.65\linewidth]{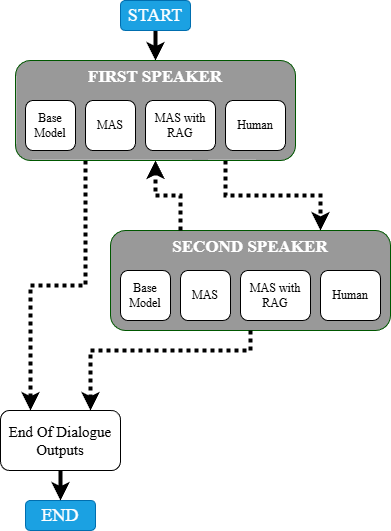}
    \caption{Parent graph controlling dialogue flow. Dashed lines indicate conditional edges determining continuation or termination.}
    \label{fig:parent-graph}
\end{figure}

We introduce \textsc{Persuasio}, a modular multi-agent dialogue platform for studying persuasive debates between humans and LLMs in UK political contexts, implemented using LangGraph.\footnote{\url{https://www.langchain.com/langgraph} (accessed: 27/2/2026)} \textsc{Persuasio} implements the formal persuasion dialogue system of \citet{10.1017/S0269888906000865} through turn-based interaction between two speakers (Fig.~\ref{fig:parent-graph}), instantiated as: (i) a base LLM agent, (ii) a structured multi-agent system (MAS), optionally augmented with retrieval-augmented generation (RAG), or (iii) a human participant. Each speaker is assigned a viewpoint to take on the political spectrum: either left-leaning or right-leaning. Dialogues terminate when a participant \textit{concedes}, \textit{retracts} their claim, or the maximum turn limit is reached.

\subsection{Dialogue Protocol}\label{subsec:dialogue-protocol}

\textsc{Persuasio} employs a multi-move, multi-reply, non-immediate reply, non-deterministic protocol, approximating real-world argumentative exchange while preserving formal tractability, allowing for rigorous logical analysis. Permissible replies (Tab.~\ref{tab:typical-replies-for-persuasio}) are adapted from \cite{10.1017/S0269888906000865} and act as constraints over generation at each turn. 
\begin{table}[h!]
\begin{adjustbox}{max width=\columnwidth}
    \centering
    \begin{tabular}{l|c}
    \toprule
        \textbf{Utterances} &  \textbf{Replies} \\
        \midrule
        \textit{claim} $\varphi$ & \textit{why} $\varphi$, \textit{claim} $\bar{\varphi}$, \textit{concede} $\varphi$, \textit{question} $\varphi$ \\
        \textit{why} $\varphi$ & $\varphi$ \textit{since} $S$, \textit{claim} $S$, \textit{retract} $\varphi$\\
        \textit{concede} $\varphi$ & \\
        \textit{retract} $\varphi$ & \\
        $\varphi$ \textit{since} $S$& \textit{why} $\psi$ ($\psi \in S$), \textit{concede} $\psi$ ($\psi \in S$), \textit{question} $\psi$ ($\psi \in S$)\\
        \textit{question} $\varphi$ & \textit{claim} $\varphi$, \textit{claim} $\bar{\varphi}$, \textit{retract} $\varphi$\\
        \bottomrule
    \end{tabular}
    \end{adjustbox}
    \caption{Typical replies for \textsc{Persuasio} {[adapted from \cite{10.1017/S0269888906000865}]}.}
    \label{tab:typical-replies-for-persuasio}
\end{table}

\subsection{Shared Architectural Components}

All agents perform \textbf{utterance classification} and \textbf{commitment updating}. The former assigns each sentence to a dialogue move using an LLM-based annotation agent with context-sensitive disambiguation (e.g., our prompts explicitly distinguish between a \textit{claim} from \textit{since} move; App.~\ref{app:subsec:prompts:utterance-classification}). Commitment updates follow the formal effect rules of \citet{10.1017/S0269888906000865}: \textit{claim} and \textit{since} add propositions; \textit{concede} adds the opponent's claim (terminating the dialogue if it is their original claim); \textit{retract} removes propositions (terminating if the speaker withdrew their own initial claim). Redundancy in commitments is resolved via LLM-based equivalence rather than string matching. Appendix~\ref{app:subsec:prompts:commitment-updates} has prompt details for commitment updates.

\subsection{Base Model Agent}
\begin{figure}[t]
    \centering
    \includegraphics[width=0.2\linewidth]{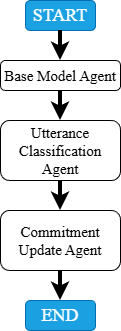}
    \caption{Base model agent.}
    \label{fig:base-model-agent}
\end{figure}

The base agent (Fig.~\ref{fig:base-model-agent}) evaluates LLM capabilities without MAS or RAG enhancements. It generates protocol-compliant responses conditioned on claims, commitments, and dialogue history (App.~\ref{app:sec:base-model-prompts}), followed by utterance classification and commitment updates.

\subsection{Structured Multi-Agent System}\label{subsec:mas}

\begin{figure}[t]
    \centering
    \includegraphics[scale=0.25]{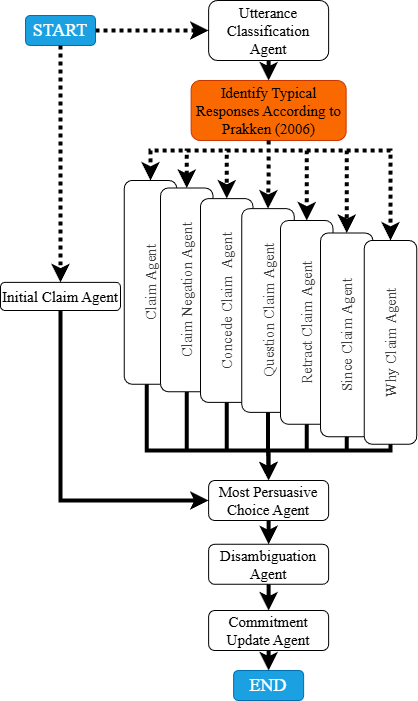}
    \caption{MAS architecture. Dashed edges indicate conditional flows; solid lines indicate unconditional traversal.}
    \label{fig:mas}
\end{figure}

The MAS (Fig.~\ref{fig:mas}) employed a divide and conquer approach to implement the persuasion dialogue system of \citet{10.1017/S0269888906000865}, decomposing response generation into the following sub-agents: a \textbf{protocol reasoner} identifies admissible replies, \textbf{generation agents} proposes candidate sentences, a \textbf{most persuasive choice agent} selects the optimal set of sentences, and a \textbf{disambiguation agent} resolves semantic overlap. Prompt details appear in Appendix~\ref{app:subsec:prompts:mas}.

\subsubsection{Graph-Based Retrieval}

\begin{figure}[t]
    \centering
    \includegraphics[scale=0.25]{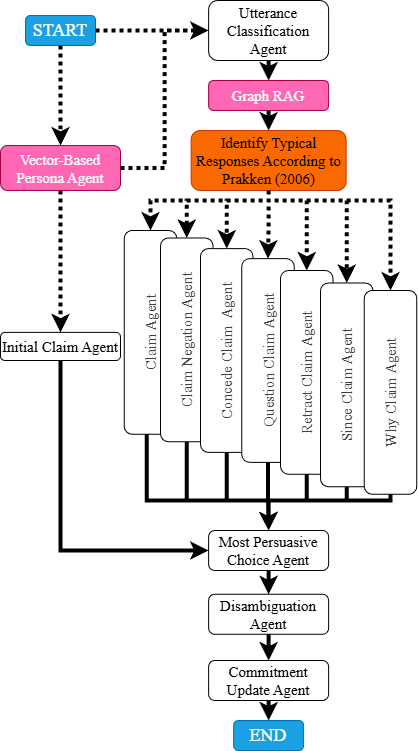}
    \caption{MAS with vector- and graph-based RAG.}
    \label{fig:mas-rag}
\end{figure}

We augment MAS with RAG (Fig.~\ref{fig:mas-rag}) using a knowledge base (KB) \cite{robinson2026validatingpoliticalpositionpredictions} containing locutions, propositions with political scores, and argumentative relations. Retrieval combines SBERT embeddings \cite{reimers-2019-sentence-bert} with cosine similarity and political alignment constraints.\footnote{Left-leaning nodes had a political score of 20--40 and right-leaning ranged from 60--80, following the 0--100 left--right scale from \cite{robinson2026validatingpoliticalpositionpredictions}.} Vector-based retrieval supplies the top-5 aligned nodes for persona construction via an LLM call (refer to App.~\ref{app:subsec:prompts:persona-gen} for prompt details); hybrid graph-based retrieval exploits the relational structure of the KB to provide pertinent examples at runtime, filtering e.g. attacks for rebuttals, supports for \textit{since} moves, etc.

\subsection{Human Sub-Graph}

\begin{figure}[h]
    \centering
    \includegraphics[width=0.3\linewidth]{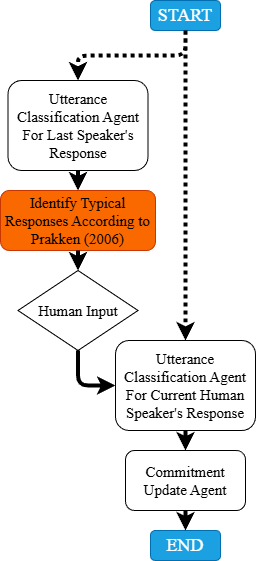}
    \caption{Human sub-graph.}
    \label{fig:human-agent}
\end{figure}

The human sub-graph follows the same formal machinery (Fig.~\ref{fig:human-agent}). The system presents admissible reply types, accepts free-text input, then applies utterance classification and commitment updates identically to artificial agents, enabling controlled cross-agent comparison.

\section{Dialogue Transcript Generation}

We generated 192 debate transcripts to evaluate persuasive performance across humans and LLMs: 144 model--model, 36 human--model, and 12 human--human dialogues. This corpus supports systematic comparison under both logical and subjective evaluation.

\paragraph{Debate Topic.}
All dialogues addressed a single UK political topic: \textit{``Debate the trade-offs of government intervention across healthcare, immigration, and welfare.''} Fixing the topic controlled for domain variation, ensuring differences in persuasiveness were attributable to interlocutor behaviour rather than subject matter.

\paragraph{Human Debaters.}
We recruited 13 undergraduate Computer Science students from the University of Liverpool. Each participated in approximately six dialogues and consented (under our ethically approved process - see page 10) to argue from both left- and right-wing perspectives. Debaters were not informed whether their opponent was human or artificial.
Opponent identities were revealed after the workshop in accordance with our ethics protocol.

\paragraph{Experimental Design.}
Interlocutors were assigned a political persona (left- or right-wing) prior to the commencement of each debate. Dialogues were started by the first speaker who asserted an initial, single-sentence claim that was based on the debate topic and aligned with their assigned political stance. The second speaker argued against their opponent's initial claim, ensuring closure under negation (see App.~\ref{app:sec:persuasion-dialogues}). Responses were constrained by the dialogue protocol (Tab.~\ref{tab:typical-replies-for-persuasio}) and limited to five sentences per turn. Dialogues that involved at least one human were time-bounded to 30 minutes; if no winner emerged within this period, a draw was recorded. Any dialogue that lasted more than 40 turns without a decisive outcome was likewise terminated and treated as a draw.

\paragraph{Models and Settings.}
Three LLMs were used in dialogues involving models (Tab.~\ref{tab:models}). To maximise reproducibility, we used near-deterministic settings (temperature=$0$, \texttt{top\_p}=$1$,\footnote{Mistral Medium required \texttt{top\_p}=$1$ when temperature=$0$.} seed=$123$). We removed the content filters from all models to ensure that they could debate freely on politically sensitive topics without refusals so as to not undermine the naturalistic quality of the exchanges. Each LLM was tested under 3 agentic configurations, i.e. \texttt{BASE} (Fig.~\ref{fig:base-model-agent}), \texttt{MAS} (Fig.~\ref{fig:mas}), and \texttt{MAS} with \texttt{RAG} (Fig.~\ref{fig:mas-rag}), yielding 9 LLM interlocutors (e.g., \texttt{GPT-4o\_BASE}, \texttt{GPT-4o\_MAS}, \texttt{GPT-4o\_MAS\_RAG}, etc), which, together with 13 human debaters, comprised the full set of $n$$=$$22$ interlocutors.
\begin{table}[h]
\centering
\begin{adjustbox}{max width=\columnwidth}
    \begin{tabular}{llcccc}
    \toprule
        \textbf{Model} & \textbf{Official Name} & \textbf{Base} & \textbf{MAS} & \textbf{MAS+RAG} & \textbf{Human Sub-Graph} \\
        \midrule
        GPT-4o & \texttt{gpt\_4o\_2024\_08\_06} & \checkmark & \checkmark & \checkmark & \checkmark \\
        Grok 3 & \texttt{grok-3} & \checkmark & \checkmark & \checkmark &  \\
        Mistral Medium & \texttt{mistral-medium-2505} & \checkmark & \checkmark & \checkmark &  \\
        \bottomrule
    \end{tabular}
\end{adjustbox}
\caption{Models used in dialogue generation.}
\label{tab:models}
\end{table}

\paragraph{Model Selection for Human Experiments.}
We selected two LLM families for the full human evaluation: GPT-4o (OpenAI) and Grok-3 (xAI). This selection was motivated by the desire to control for ideological diversity that has the potential to be present in the underlying foundation models. Prior work has demonstrated that OpenAI's ChatGPT exhibits measurable left-leaning tendencies \cite{motoki2024more}. As different model providers employ distinct training data pipelines and alignment strategies, pairing models from OpenAI and xAI allowed us to reduce the risk that observed rankings of persuasiveness were artefacts of a single provider's alignment regime rather than of the agentic or model architecture itself.

\paragraph{Utterance Classification Evaluation.}
The human sub-graph requires accurate dialogue move classification and commitment updating. We evaluated classification performance on a manually annotated test set drawn from the KB of \citet{robinson2026validatingpoliticalpositionpredictions}. Using the same settings (temperature=$0$, \texttt{top\_p}=$1$, seed=$123$), we computed macro-F1 with 10{,}000 bootstrap resamples and report 95\% confidence intervals in Appendix~\ref{app:utt-class-eval}. GPT-4o achieved the strongest performance and was therefore used for classification and commitment updates in the human sub-graph for dialogues with at least one human interlocutor.

\section{Quantifying Logical and Subjective Persuasiveness}\label{sec:quantifying-persuasiveness}

To derive both logical and subjective accounts of persuasiveness across human and model debaters, we fit BT models to two complementary data sources:

\begin{itemize}[parsep=0pt]
\item \textbf{Debate outcomes}: Win--loss records from human--human, human--model, and model--model debates conducted on the \textsc{Persuasio} platform; and
\item \textbf{Pairwise transcript judgements}: Comparative annotations elicited from crowdworkers recruited via Prolific,\footnote{\url{https://www.prolific.com} (accessed: 27/2/2026)} who evaluated pairs of dialogue excerpts for persuasiveness. All participants resided in the UK, reported English as their first or primary language, and had completed some form of higher education.
\end{itemize}

\paragraph{Debate Outcomes.}
\textsc{Persuasio} adjudicated a winner for each debate at runtime based on the formal properties of the dialogue. Across the 192 debates conducted on the platform, these outcomes were recorded and used to construct a logical, performance-based ranking of the interlocutors.

\paragraph{Sampling Transcript Pairs.}
We sampled from the pool of 192 debate transcripts generated using \textsc{Persuasio}. Across all debates, there were 22 distinct interlocutors -- 9 model variants and 13 humans -- each appearing in at least one debate, yielding $\binom{22}{2} = 231$ unique interlocutor pairings. Each annotation instance contained two excerpts drawn from different debates. To ensure comparability, we constrained both excerpts such that Speaker~A adopted the same ideological position in each (e.g., left-wing in both). Excerpts consisted of four consecutive dialogue turns. To ensure balanced coverage, for each of the 231 pairings we sampled six excerpt pairs without replacement: three in which Speaker~A argued from a left-wing perspective and three from a right-wing perspective,\footnote{We drew six left-wing samples for \texttt{Human\_2}, as this participant did not take part in any right-wing debates.} yielding a total of 1,386 annotation instances.

\paragraph{Reliable BT Rankings.}
Both evaluation sources exceed the $n \ln n \approx 68$ pairwise comparisons sufficient for reliable ranking recovery under the BT model~\cite{negahban2012_ranking}, where $n$$=$$22$ in our setting. The debate-outcome matrix contained 192 comparisons, and the human-annotation matrix included 1{,}386.

\paragraph{Human Annotation of Debate Excerpts.}
Each of the 1,386 samples were independently evaluated by 7 crowdworkers (3,234 unique annotators in total). For each instance, annotators compared the persuasiveness of Speaker~A across the two excerpts, selecting which was more persuasive or indicating that both were equally persuasive. Annotators additionally reported their confidence on a five-point Likert scale~\cite{likert1932technique}, ranging from ``Very unsure'' to ``Very confident''. Each annotator completed three comparisons, responding to two questions per sample:
\begin{enumerate}[parsep=0pt]
\item \emph{``Based only on the debate transcripts, which Speaker~A presented the more persuasive arguments?''}
\item \emph{``How confident are you in your choice?''}
\end{enumerate}

\paragraph{Win Matrices.}
We constructed two $n \times n$ win matrices ($n$$=$$22$). In each matrix, entry $W_{ij}$ recorded the number of times interlocutor~$i$ was judged more persuasive than interlocutor~$j$. The \emph{logical} win matrix was derived from automated debate adjudication: $W_{ij}$ was incremented by $1$ whenever interlocutor~$i$ won a debate against~$j$. Ties were split equally, incrementing both $W_{ij}$ and $W_{ji}$ by $0.5$~\cite{davidson1970}. This matrix was sparse, as not all interlocutors debated one another and no interlocutor debated itself. The \emph{subjective} win matrix was derived from crowdsourced annotations: for each comparison, $W_{ij}$ was incremented by the number of annotators who preferred interlocutor~$i$, while $W_{ji}$ was incremented by the count favouring~$j$. By design, all $\binom{22}{2}$ unique pairings were sampled, each receiving 42 annotations, ensuring every off-diagonal entry was non-zero.

\paragraph{Pairwise Modelling.}
We fitted BT models to each win matrix using the Iterative Luce Spectral Ranking (I-LSR) algorithm~\cite{maystre2015_ilsr}, implemented in the Python library \texttt{choix},\footnote{\url{https://choix.lum.li/} (accessed: 27/2/2026)} with regularisation parameter $\alpha = 0.01$. This yielded latent strength parameters $\boldsymbol{\theta} \in \mathbb{R}^n$, where the probability of interlocutor~$i$ being preferred over~$j$ is given by $p(i \succ j) = \frac{e^{\theta_i}}{e^{\theta_i} + e^{\theta_j}}$. Rankings were obtained by sorting interlocutors in descending order of $\theta_i$, with higher values indicating greater estimated persuasiveness. Applying this procedure to the subjective win matrix produced the subjective ranking $\mathcal{R}^{(\text{subj})}$; applying it to the logical win matrix produced the logical ranking $\mathcal{R}^{(\text{logi})}$. We additionally fitted subgroup-specific rankings $\mathcal{R}^{(g)}$ by partitioning the subjective annotations according to self-declared demographic attributes (age, gender, and political leaning), constructing separate win matrices for each subgroup, and applying the same BT/I-LSR procedure.

\paragraph{Evaluation.}
We quantified distance and agreement between rankings over the $n$$=$$22$ interlocutors using three metrics: Kendall's $\tau$ distance~\cite{kendalltau}, normalised to $[0,1]$ by dividing by its maximum $\binom{n}{2}$ (denoted $d_{\tau}$); Spearman's footrule distance~\cite{spearmanfr}, normalised by $\lfloor n^2/2 \rfloor$ (denoted $d_\text{footrule}$); and ordinal Krippendorff's $\alpha$~\cite{Hayes01042007,krippendorff2018content}. We first report these metrics between the global subjective and logical rankings ($\mathcal{R}^{(\text{subj})}$ vs.\ $\mathcal{R}^{(\text{logi})}$) to quantify their overall divergence. We then compare each subgroup ranking $\mathcal{R}^{(g)}$ to both global rankings, allowing us to examine how closely different demographic groups align with subjective and logical persuasiveness.

\section{Experimental Evaluation}
\label{sec:experimental_evaluation}

We now examine the relationship between two conceptions of persuasiveness across $n$$=$$22$ interlocutors: \emph{logical} persuasiveness, derived from automated adjudication grounded in formal argumentation theory \cite{10.1017/S0269888906000865}, and \emph{subjective} persuasiveness, derived from crowdsourced pairwise judgements of rhetorical appeal. Our central question is whether these rankings coincide: do those who win on logical grounds also appear most persuasive to human evaluators?

\paragraph{Inter-Rater Reliability.}
Nominal Krippendorff's $\alpha$ was $0.188$ for persuasiveness and $0.04$ for confidence annotations. Although these values fall below conventional thresholds for strong agreement, low agreement is expected in annotation tasks involving inherently subjective constructs such as persuasiveness \cite{passonneau-carpenter-2014-benefits}. Evaluations of rhetorical strength are shaped by prior beliefs, stylistic preferences, and individual standards of argument quality. Rather than reflecting annotation noise, the observed disagreement underscores the irreducibly subjective nature of the construct. As we show below, aggregate rankings remain highly stable despite item-level variability.

\subsection{Subjective vs.\ Logical Persuasiveness}
Table~\ref{tab:persuasiveness_comparison} presents the subjective and logical persuasiveness rankings for all $n$$=$$22$ interlocutors.
\begin{table}[h]
\centering
\begin{adjustbox}{max width=\columnwidth}
\begin{tabular}{c lc lc}
\toprule
 & \multicolumn{2}{c}{$\mathcal{R}^{(\text{subj})}$} & \multicolumn{2}{c}{$\mathcal{R}^{(\text{logi})}$} \\
\cmidrule(lr){2-3} \cmidrule(lr){4-5}
Rank & Interlocutor & $\hat{\theta}_i$ & Interlocutor & $\hat{\theta}_i$ \\
\midrule
1 & \texttt{GPT-4o\_MAS\_RAG} & 1.000 & \texttt{Human\_10} & 1.000 \\
2 & \texttt{Grok-3\_MAS\_RAG} & 0.988 & \texttt{Mistral-Medium\_BASE} & 0.817 \\
3 & \texttt{GPT-4o\_BASE} & 0.956 & \texttt{Grok-3\_MAS} & 0.774 \\
4 & \texttt{Grok-3\_MAS} & 0.920 & \texttt{GPT-4o\_BASE} & 0.760 \\
5 & \texttt{GPT-4o\_MAS} & 0.863 & \texttt{Mistral-Medium\_MAS\_RAG} & 0.747 \\
6 & \texttt{Mistral-Medium\_MAS\_RAG} & 0.850 & \texttt{Human\_2} & 0.737 \\
7 & \texttt{Mistral-Medium\_MAS} & 0.774 & \texttt{Human\_8} & 0.705 \\
8 & \texttt{Mistral-Medium\_BASE} & 0.764 & \texttt{Mistral-Medium\_MAS} & 0.701 \\
9 & \texttt{Grok-3\_BASE} & 0.742 & \texttt{Human\_5} & 0.701 \\
10 & \texttt{Human\_4} & 0.654 & \texttt{Human\_9} & 0.699 \\
11 & \texttt{Human\_7} & 0.641 & \texttt{GPT-4o\_MAS} & 0.694 \\
12 & \texttt{Human\_8} & 0.637 & \texttt{Grok-3\_BASE} & 0.639 \\
13 & \texttt{Human\_5} & 0.605 & \texttt{GPT-4o\_MAS\_RAG} & 0.616 \\
14 & \texttt{Human\_2} & 0.536 & \texttt{Grok-3\_MAS\_RAG} & 0.568 \\
15 & \texttt{Human\_10} & 0.506 & \texttt{Human\_12} & 0.546 \\
16 & \texttt{Human\_13} & 0.467 & \texttt{Human\_6} & 0.528 \\
17 & \texttt{Human\_11} & 0.377 & \texttt{Human\_13} & 0.520 \\
18 & \texttt{Human\_6} & 0.364 & \texttt{Human\_1} & 0.519 \\
19 & \texttt{Human\_3} & 0.333 & \texttt{Human\_11} & 0.516 \\
20 & \texttt{Human\_12} & 0.316 & \texttt{Human\_4} & 0.451 \\
21 & \texttt{Human\_1} & 0.212 & \texttt{Human\_7} & 0.415 \\
22 & \texttt{Human\_9} & 0.000 & \texttt{Human\_3} & 0.000 \\
\bottomrule
\end{tabular}
\end{adjustbox}
\caption{Subjective ($\mathcal{R}^{(\text{subj})}$) and logical ($\mathcal{R}^{(\text{logi})}$) persuasiveness rankings of $n$$=$$22$ interlocutors, with min-max normalised BT strength scores $\hat{\theta}_i \in [0,1]$.}
\label{tab:persuasiveness_comparison}
\end{table}

\paragraph{LLM Dominance in Subjective Persuasion.}
All 9 LLM configurations occupied the top nine positions in the subjective ranking. The highest-ranked human (\texttt{Human\_4}) appeared only at rank~10, separated from the lowest-ranked LLM (\texttt{Grok-3\_BASE}) by a non-trivial margin. Subjectively, model-generated excerpts were consistently judged more persuasive than those produced by humans.

\paragraph{Human Competitiveness in Logical Persuasion.}
The logical ranking exhibited a markedly different structure. A human debater (\texttt{Human\_10}) came top, and humans occupied six of the top ten ranks. Unlike the subjective ranking, logical persuasiveness was not dominated by the agentic variants; instead, humans competed throughout the upper tier.

\paragraph{Systematic Rank Inversions.}
The most striking pattern is the presence of substantial rank inversions between paradigms. The two highest-ranked subjective agentic systems (\texttt{GPT-4o\_MAS\_RAG} and \texttt{Grok-3\_MAS\_RAG}) dropped to mid-table positions under logical evaluation. Conversely, some systems with modest subjective standing (i.e., \texttt{Mistral-Medium\_BASE}) rose sharply under logical criteria. These inversions demonstrate that rhetorical appeal and logical soundness are not interchangeable dimensions of performance.

\paragraph{Effect of MAS and RAG Enhancements.}
MAS and RAG configurations consistently improved subjective persuasiveness, often elevating models like GPT-4o and Grok-3 to the top of the ranking. However, their impact on logical persuasiveness was inconsistent and, in several cases, negative. Enhancements that amplify fluency, structure, and evidential density appear to increase rhetorical impact without reliably improving inferential coherence.

\paragraph{Human Variability.}
Human debaters exhibited substantially greater variance across both rankings. The strongest humans matched or surpassed LLMs in logical persuasiveness, while the weakest ranked near the bottom of both scales. In contrast, LLM configurations were tightly clustered, particularly in the subjective ranking, indicating more homogeneous performance profiles.

\subsection{Demographic Stability of Persuasiveness Rankings}

Table~\ref{tab:subgroup_ranking_agreement} reports ranking distance and agreement between demographic subgroup rankings and the global subjective and logical rankings.
\begin{table}[htbp]
\centering
\begin{adjustbox}{max width=\columnwidth}
\begin{tabular}{l cc cc cc cc}
\toprule
 & & & \multicolumn{2}{c}{$d_{\tau}$} & \multicolumn{2}{c}{$d_{\text{footrule}}$} & \multicolumn{2}{c}{Kripp.\ $\alpha$} \\
\cmidrule(lr){4-5} \cmidrule(lr){6-7} \cmidrule(lr){8-9}
 & $n$ & $|$Ann.$|$ & $\mathcal{R}^{(\text{subj})}$ & $\mathcal{R}^{(\text{logi})}$ & $\mathcal{R}^{(\text{subj})}$ & $\mathcal{R}^{(\text{logi})}$ & $\mathcal{R}^{(\text{subj})}$ & $\mathcal{R}^{(\text{logi})}$ \\
\midrule
$\mathcal{R}^{(\text{subj})}$ vs.\ $\mathcal{R}^{(\text{logi})}$ & 3234 & 9702 & \multicolumn{2}{c}{0.381} & \multicolumn{2}{c}{0.504} & \multicolumn{2}{c}{0.410} \\
\midrule
Left-Leaning & 1591 & 4773 & 0.030 & 0.377 & 0.058 & 0.496 & 0.990 & 0.427 \\
Centre-Leaning & 1194 & 3582 & 0.035 & 0.381 & 0.058 & 0.488 & 0.988 & 0.422 \\
Right-Leaning & 447 & 1341 & 0.108 & \textbf{0.299} & 0.174 & \textbf{0.397} & 0.922 & \textbf{0.614} \\
\midrule
Age 18--30 & 774 & 2322 & 0.043 & 0.390 & 0.074 & 0.545 & 0.983 & 0.374 \\
Age 31--40 & 981 & 2943 & 0.056 & 0.359 & 0.099 & 0.455 & 0.977 & 0.491 \\
Age 41--50 & 679 & 2037 & 0.078 & 0.355 & 0.140 & 0.488 & 0.957 & 0.463 \\
Age 51--60 & 475 & 1425 & 0.082 & 0.359 & 0.124 & 0.446 & 0.957 & 0.506 \\
Age 61--70 & 265 & 795 & 0.104 & 0.433 & 0.157 & 0.537 & 0.934 & 0.272 \\
Age 70+ & 58 & 174 & 0.221 & 0.403 & 0.306 & 0.545 & 0.776 & 0.315 \\
\midrule
Female & 1860 & 5580 & \textbf{0.026} & 0.390 & \textbf{0.050} & 0.521 & \textbf{0.991} & 0.380 \\
Male & 1365 & 4095 & 0.048 & 0.351 & 0.091 & 0.438 & 0.980 & 0.480 \\
\bottomrule
\end{tabular}
\end{adjustbox}
\caption{Ranking agreement between each demographic subgroup ranking $\mathcal{R}^{(g)}$ and the global subjective ($\mathcal{R}^{(\text{subj})}$) and logical ($\mathcal{R}^{(\text{logi})}$) rankings, where $n$ denotes the number of unique annotators in each subgroup. Lower $d_{\tau}$ and $d_{\text{footrule}}$ indicate closer agreement; higher $\alpha$ indicates stronger agreement. Best subgroup values per column are in bold.}
\label{tab:subgroup_ranking_agreement}
\end{table}

\paragraph{Global Divergence.}
Agreement between the global subjective and logical rankings was moderate (ordinal $\alpha = 0.410$), while distance metrics indicated substantial positional shifts. This confirms that the two rankings capture overlapping but distinct aspects of persuasion.

\paragraph{Robustness of Subjective Judgements.}
All demographic subgroups closely reproduced the global subjective ranking, with near-perfect agreement across political, age, and gender categories. Despite low item-level inter-rater reliability, aggregate subjective rankings are remarkably stable. This supports the use of pairwise modelling for inherently subjective constructs: individual judgements vary, yet their aggregate structure is consistent.

\paragraph{Divergence from Logical Ranking.}
In contrast, every demographic subgroup diverged substantially from the logical ranking. Human perceptions of persuasiveness across demographic subgroups systematically differed from formal measures of persuasive argumentation.

\paragraph{Political Leaning.}
Right-leaning annotators constituted a notable exception. While they deviated most from the global subjective consensus, they aligned more closely with the logical ranking. This suggests that this subgroup may place relatively greater persuasive weight on certain aspects of argumentation, such as logic, rationale and chain of reasoning, when compared to rhetorical, linguistic and stylistic features of debate transcripts. However, given subgroup size differences, this finding should be interpreted cautiously.

\paragraph{Age and Gender.}
Agreement with the subjective ranking declined with age, though without corresponding convergence toward the logical ranking. Older annotators appear to apply evaluative criteria distinct from both paradigms studied here. Gender differences were modest: female annotators aligned most closely with the global subjective ranking, whereas male annotators showed slightly greater alignment with the logical ranking. Since effect sizes are small, this asymmetry should also be treated with caution.

\subsection{Discussion}

Across analyses, a consistent divide emerges: LLMs dominate subjective persuasiveness yet fall sharply in logical ranking, while humans exhibit the opposite pattern. The LLMs and agentic variants studied appear optimised for rhetorical polish -- fluency, structural clarity, confident framing -- rather than formal argumentative rigour. Logical adjudication, grounded in formal coherence, rewards different properties than those that drive human perceptions of persuasiveness.

The pronounced rank inversions among top-performing subjective systems perhaps highlights a fundamental limitation of LLMs for logical reasoning: their stochastic, autoregressive paradigm renders them inherently myopic, precluding strategic planning or deliberate reasoning over upcoming content. This limitation is potentially compounded by post-training procedures -- namely instruction tuning and reinforcement learning from human feedback ~\cite{ouyang2022traininglanguagemodelsfollow} -- which optimise for rhetorical fluency and may thereby further obscure deficiencies in logical reasoning. Our results show that multi-agent orchestration and retrieval augmentation seemingly amplify this effect by enhancing perceived persuasiveness without reliably safeguarding logical integrity.

\paragraph{Implications for Evaluation and Deployment.}
These findings carry implications for model evaluation and deployment. Benchmarks relying solely on human preference risk conflating fluency with soundness. In high-stakes domains that require critical thought -- such as legal reasoning, public policy, and education -- optimising for perceived persuasiveness alone may produce discourse that is compelling yet logically brittle. Aligning rhetorical strength with argumentative validity may require explicit architectural constraints, such as structured reasoning verification, claim--evidence consistency checks, or argumentation-theoretic supervision during LLM training or prompt engineering.


\section{Conclusions and Future Work}
\label{sec:conclusion}

We introduced \textsc{Persuasio}, a modular multi-agent dialogue platform grounded in the formal, argumentation-based persuasion dialogue system of \citet{10.1017/S0269888906000865}, enabling controlled argumentative exchanges between humans and LLMs in the context of UK political discourse. Using \textsc{Persuasio}, we collected 192 debate transcripts across human--human, human--model, and model--model settings, and evaluated 22 interlocutors along two complementary axes: (i) \emph{logical persuasiveness}, derived from automated dialogue adjudication, and (ii) \emph{subjective persuasiveness}, based on 9{,}702 crowdsourced pairwise comparisons.

Our main result is a systematic separation between these dimensions. All nine LLM configurations occupied the top nine positions in the subjective ranking, yet saw substantial rank inversions under logical evaluation, where human debaters were much more competitive. This divergence suggests that the rhetorical polish that makes LLM-generated arguments subjectively compelling does not reliably translate into logical soundness. Optimising for fluency, organisation, and assertive presentation does not reliably yield arguments that are logically consistent or well-supported by evidence.

These findings raise important considerations for the evaluation and deployment of LLMs in deliberative domains such as legal reasoning, public policy, etc. Evaluation protocols grounded exclusively in human preference risk conflating persuasive appeal with argumentative validity, thereby incentivising systems that optimise for surface-level persuasiveness rather than substantive reasoning quality.


Future work should explore aligning these dimensions through explicit logical quality signals in prompting or post-training pipelines, and through hybrid human-AI systems that combine LLM fluency with human logical scrutiny. The demographic patterns observed also merit replication with larger, more balanced samples.

\newpage

\section*{Limitations}
\label{sec:limitations}

We acknowledge several limitations that should be considered when interpreting our findings.

\paragraph{Exploratory Design.}
Our evaluation is exploratory rather than confirmatory: we do not test pre-registered hypotheses, and we do not report $p$-values, confidence intervals, or compare observed rank distances against null distributions. We report multiple rank-distance and agreement metrics (Kendall's $\tau$ distance, Spearman's Footrule distance, and Krippendorff's $\alpha$) to characterise the relationship between subjective and logical persuasiveness rankings, but we cannot determine whether the observed agreements or disagreements are statistically significant or fall within the range expected by chance. 

\paragraph{Small Ranking Space.}
Our rankings comprise only $n$$=$$22$ interlocutors (9~LLM configurations and 13~humans), which limits the discriminative power of rank-based metrics. With so few items, normalised distances are coarse-grained, and small perturbations in the underlying win matrices can produce notable shifts in rank ordering. This makes it difficult to draw fine-grained conclusions about relative persuasiveness, particularly among interlocutors with similar BT strength estimates.

\paragraph{Demographic Subgroup Analysis.}
The subgroup analyses partition the annotation pool by self-declared age, gender, and political leaning, yielding groups of varying and sometimes small sizes (e.g., $174$ annotations across 58 annotators aged 70 or older; $1{,}341$ annotations from 447 right-leaning annotators). Subgroup-specific rankings derived from fewer annotations are inherently less stable, and the observed variation in alignment with global rankings may partly reflect differences in sample size rather than genuine demographic effects. We do not apply multiple-comparison corrections across the eleven subgroups examined, increasing the risk that apparent patterns arise from sampling variability. The demographic patterns surfaced here, particularly the right-leaning and age-related effects, should be treated as hypotheses for confirmatory study with larger, balanced samples rather than as definitive claims.

\paragraph{Annotator Population.}
Crowdsourced annotations were collected via Prolific and are subject to the demographic and attitudinal biases of that platform's participant pool. The sample may not be representative of the broader population, and subjective persuasiveness judgements are likely influenced by cultural, linguistic, and educational factors that we do not control for. Annotator engagement and attention may also vary across tasks, potentially introducing noise into the subjective win matrix. Additionally, some human debaters reported unfamiliarity with the distinction between left- and right-wing political positions, which may have introduced noise into the debate transcripts and, by extension, the annotation data. Using a shared initial claim rather than ideological labels might have mitigated this issue.

\paragraph{Logical Persuasiveness.}
The logical ranking $\mathcal{R}^{(\text{logi})}$ is derived from \textsc{Persuasio}, underpinned by the formalism of \citet{10.1017/S0269888906000865}, in which each LLM participant is itself responsible for judging whether it has lost the debate. Because \textsc{Persuasio} treats each debater as an autonomous agent, adjudication is not performed by a separate, independent model; instead, the outcome is determined by whether an LLM engaged in the debate believes it asserted a \textit{concede} or \textit{retract} move that should lead to defeat. We adopted this self-adjudication design on the assumption that it most closely mirrors how humans engage in debate -- i.e., a participant recognises when their position has been undermined rather than deferring to an external arbiter. However, this choice introduces potential biases: different LLM architectures may vary systematically in their willingness to concede, and some models may be overly deferential or resistant to acknowledging defeat, distorting the resulting win matrix. A separate, neutral adjudicator model might yield different and potentially more consistent outcomes, and future work should compare self-adjudicated results against independent third-party evaluation. Additionally, the logical win matrix is sparse -- i.e. not all interlocutor debate pairings were generated between models and humans -- which may bias BT estimates toward interlocutors with less frequently observed matchups. Notably, \texttt{Mistral-Medium\_BASE}'s strong logical ranking should be interpreted cautiously given its absence from any debates with human interlocutors.

\paragraph{BT Model Assumptions and Tie Handling.}
The BT model assumes pairwise outcomes are governed by a single latent strength parameter per interlocutor, implying transitivity and context-independence. In practice, persuasiveness may be topic-, opponent-, or context-dependent. The regularisation parameter $\alpha = 0.01$ mitigates convergence issues for sparse matrices but introduces a prior that shrinks estimates toward uniformity, potentially obscuring genuine differences among interlocutors with limited comparison data. Draws were resolved by assigning half-wins to each party, a standard convention~\cite{davidson1970} that assumes ties are uninformative about relative strength, an assumption that may not hold when ties arise from annotator uncertainty or adjudication ambiguity rather than genuinely equal ability.

\paragraph{Generalisability.}
Our findings are specific to the debate topic, interlocutor configurations, and evaluation protocols used in this study. All dialogues addressed a single political topic grounded in current UK discourse; the extent to which the observed relationships between logical and subjective persuasiveness generalise to other domains, languages, debate formats, agentic sub-graphs, or LLM architectures remains an open question. The crowdsourced annotations capture a single snapshot of perceived persuasiveness; longitudinal or deliberative designs, in which annotators engage with arguments over multiple rounds, may yield different patterns.

\section*{Ethical Considerations}

This study received ethical approval from the University of Liverpool (Ref: 17359), the University of Leeds (Ref: MEEC 25-001), and the Alan Turing Institute (Ref: TR25-17).
The removal of content filters from all models used for debate transcript generation was reviewed and approved as part of this process, on the grounds that it was necessary to ensure a fair comparison of human and model persuasiveness. The choice of political debate topic was likewise deliberate: recruiting participants from within the UK ensured that all debaters possessed sufficient familiarity with the subject matter to engage substantively.

\paragraph{Human Debaters.} Thirteen human debaters were recruited from the University of Liverpool and participated in generating the human--model and human--human debate transcripts. They were compensated at a rate of £18.60 per hour for a total of five hours of participation. All debaters provided informed consent prior to taking part. To preserve the validity of the debates, participants were not informed of their opponent's identity (i.e. human or LLM-based agent) during the course of each dialogue, and identities were disclosed upon conclusion, in accordance with our approved ethics protocol. Participants were also explicitly informed that they could withdraw from the platform at any time, including if they found any exchange distressing. No participant exercised this option during the workshop. To ensure that no personally identifiable information was included in the accompanying released resource, each debater was assigned a unique anonymous identifier (e.g., \texttt{Human\_1}--\texttt{Human\_13}).

\paragraph{Crowdsourced Annotators.} Annotators were recruited through the online research platform Prolific to evaluate the persuasiveness of debate transcripts. All participants provided informed consent prior to undertaking the annotation tasks, in accordance with institutional ethical guidelines. Prior to the annotation tasks, we checked all annotation samples (i.e., the dialogue excerpts) presented to annotators for offensive content, and found there to be none. No personally identifiable information was collected by the research team; participant anonymisation was managed entirely by the Prolific platform. Annotators were compensated at an average rate of £11.49 per hour, above the platform's recommended minimum rate. The annotation guidelines, including task instructions and attention-check procedures, are provided in Appendix~\ref{app:subsec:annotation-guidelines}.

\section*{Data and Code Availability}


Our code, prompts and instructions for reproducing all experiments has been made publicly available: \href{https://github.com/RobinsonJI/Evaluating-the-Capabilities-of-LLMs-for-Persuasive-Dialogue}{https://github.com/RobinsonJI/Evaluating-the-Capabilities-of-LLMs-for-Persuasive-Dialogue}. All data and code are released under the MIT License.

\section*{Acknowledgements}
The work reported in this paper was supported by funding from The Alan Turing Institute. 

\bibliography{refs}

\appendix

\section{Argumentation-Based Persuasion Dialogues}\label{app:sec:persuasion-dialogues}

We ground our system in the dialogue-theoretic tradition of formal argumentation, adopting and operationalising the persuasion dialogue framework of \cite{10.1017/S0269888906000865}. Our work is situated within the broader typology of dialogue types introduced by \citep{walton1995commitment}, where argumentation-based dialogues are categorised into six types: persuasion, negotiation, information-seeking, deliberation, inquiry, and eristic. Persuasion dialogues are characterised by an initial conflict of opinions and a goal of resolving this conflict through regulated argumentative discourse. 

Persuasion dialogue systems are governed by the \textit{communication language} and \textit{dialogue protocol}. The communication language constrains what participants may utter within a debate. The protocol determines when dialogue moves are legal, how they affect the dialogue state, and when the dialogue terminates.

Following \citeauthor{10.1017/S0269888906000865} (\citeyear{10.1017/S0269888906000865}), a persuasion dialogue system is comprised of:
\begin{itemize}
    \item \textbf{Dialogue goal}: Resolution of a conflict between participants.
    \item \textbf{Topic language $\mathcal{L}_t$}, closed under classical negation.
    \item \textbf{Communication language $\mathcal{L}_c$}, containing the following locutions:
        \begin{itemize}
            \item \textit{claim} $\varphi$: assert that $\varphi$ holds;
            \item \textit{why} $\varphi$: challenge $\varphi$ by requesting justification;
            \item \textit{concede} $\varphi$: accept $\varphi$ is the case;
            \item \textit{retract} $\varphi$: withdraw commitment to $\varphi$\footnote{A \textit{retract} $\varphi$ dialogue move can only be made if the speaker is committed to $\varphi$, otherwise it is a proclamation of non-commitment, e.g. in response to a question. };
            \item $\varphi$ \textit{since} $S$: assert $\varphi$ is supported by premises $S$;
            \item \textit{question} $\varphi$: request another participant's stance on $\varphi$.
        \end{itemize}
    \item \textbf{Protocol}, specifying legal replies, turn-taking structure and termination conditions.
    \item \textbf{Participants}, each associated with a role (i.e., proponent, opponent or neutral towards a topic), internal beliefs and values, and publicly accessible commitments. 
    \item \textbf{Effect rules}, which define how individual speech acts change the state of the dialogue commitments. 
    \item \textbf{Outcome rules}, defining the winners and losers of the persuasion dialogue.
\end{itemize}

Commitments play a central role in persuasion dialogue systems: they represent publicly undertaken propositions and enforce dialogical consistency. Dialogue progression and termination are defined in terms of structured transformations over the set of commitments, as well as an interlocutor's initial claim. An interlocutor wins a debate if their opponent makes a \textit{concede} move that agrees with their original claim. Alternatively, the dialogue also terminates if the opponent makes a \textit{retract} move that withdraws their original claim in favour of the interlocutor's stance.

\section{Prompts}\label{app:sec:prompts}

For reproducibility, we summarise the prompt design used in our experiments. All prompts are fully integrated into the released codebase. 

\paragraph{Formatting and Output.}
All agents were instructed to return JSON-only outputs with no explanatory text, ensuring machine-readable responses for downstream components.

\subsection{Utterance Classification}\label{app:subsec:prompts:utterance-classification}

\textsc{Persuasio} models dialogue interpretation as a sentence-level classification task over six dialogue moves (Tab.~\ref{tab:typical-replies-for-persuasio}). An LLM acts as an expert annotator. The system prompt defines each move type and enforces deterministic, mutually exclusive, single-label assignment. Prior turns are embedded using XML-style tags (e.g., \texttt{<claim>}, \texttt{<why>}, \texttt{<since>}) to enable context-sensitive reasoning.

\subsubsection{Utterance Type Definitions}

Classification conditions on the two preceding dialogue turns to disambiguate context-dependent moves (e.g., distinguishing \emph{claim} from \emph{since}). Interrogatives are split into justification-seeking (\emph{why}) and neutral information/opinion requests (\emph{question}). Explicit agreement is \emph{concede}; explicit withdrawal is \emph{retract}. The categories are:

\begin{itemize}
    \item \textbf{claim}: asserts proposition $\phi$.
    \item \textbf{since}: asserts $\phi$ and provides supporting reasons in response to a prior challenge.
    \item \textbf{why}: challenges $\phi$ and requests justification.
    \item \textbf{question}: requests opinion or information about $\phi$ (non-justificatory).
    \item \textbf{concede}: explicitly agrees with a prior $\phi$.
    \item \textbf{retract}: explicitly withdraws a prior $\phi$.
\end{itemize}

\subsubsection{Classification Rules}

The prompt enforces:

\begin{enumerate}
    \item Exactly one label per sentence.
    \item Attacks are classified as \emph{claim}.
    \item Reason-giving assertions are \emph{since} only if preceded by a challenge; otherwise \emph{claim}.
    \item Context is required to distinguish \emph{since} from \emph{claim}.
    \item Justification-seeking challenges are \emph{why}.
    \item Sentences containing ``why'' are \emph{why}; other interrogatives are \emph{question}.
    \item \emph{Question} is restricted to neutral information/opinion requests.
    \item Explicit agreement is \emph{concede}.
    \item Explicit self-withdrawal is \emph{retract}.
\end{enumerate}

Typed moves feed downstream protocol enforcement and commitment updates.

\subsection{Commitment Updates}\label{app:subsec:prompts:commitment-updates}

Commitments are dynamically maintained sets of propositions for each participant, updated after each turn according to the formal effect rules of \citet{10.1017/S0269888906000865}. Each sentence (with an assigned move type) is processed sequentially from the current speaker's perspective.

\paragraph{Move Effects.}
\begin{itemize}
    \item \textbf{\emph{claim} $\varphi$}: add $\varphi$ to the speaker's commitments if absent; the first such claim is marked as the \emph{initial claim}.
    \item \textbf{\emph{since} $\psi$}: add supporting premises if absent.
    \item \textbf{\emph{concede} $\varphi$}: add $\varphi$; conceding the opponent's initial claim terminates the dialogue (opponent wins).
    \item \textbf{\emph{retract} $\varphi$}: remove $\varphi$; retracting one's initial claim terminates the dialogue.
\end{itemize}

To prevent duplication, semantic equivalence is determined via LLM-based judgements rather than string or embedding matching.

\subsubsection{Prompt Architecture}\label{app:subsubsec:prompts:commitment-update-details}

Each update operation is a separate LLM call with a fixed system prompt (defining the role as a \emph{commitment update manager}) and a dynamic user prompt containing the relevant dialogue state. 

\paragraph{Adding \emph{claim} and \emph{since}.}
Before insertion, a semantic redundancy check determines whether the proposition is already entailed by the commitment set. If not, it is appended; otherwise discarded.

\paragraph{Semantic Redundancy Check.}
Given the commitment list and candidate proposition, the model answers:
\begin{quote}
\textit{Is the utterance already in or semantically similar to the user's set of commitments?}
\end{quote}
Output: \texttt{\{``similar'': true/false\}}.

\paragraph{Concession Checks.}
For \emph{concede} moves, the model checks agreement with any opponent commitment:
\begin{quote}
\textit{Does the utterance concede to a claim within the user's set of commitments?}
\end{quote}
Output: \texttt{\{``concede'': true/false\}}.

A separate prompt tests concession to the opponent's \emph{initial claim}, which triggers dialogue termination.

\paragraph{Retraction Checks.}
Retraction of the speaker's initial claim is tested separately:
\begin{quote}
\textit{Does the user's last utterance retract their initial claim?}
\end{quote}
Output: \texttt{\{``retract'': true/false\}}.

General retractions are detected against the full commitment set:
\begin{quote}
\textit{Does the user's last utterance retract any of their commitments?}
\end{quote}

If true, a follow-up prompt identifies the exact retracted propositions, returning:
\texttt{\{"retracted\_sentences": [...] \}}. The retracted propositions are then removed from the speaker's commitments if present.

\subsubsection{Pipeline Summary}\label{app:subsubsec:prompts:commitment-summary}

All commitment operations follow a uniform template: a static system role definition paired with a dynamically constructed user message embedding the current dialogue state. This modular design enables independent evaluation of each formal effect rule while maintaining structured, parseable outputs.

\subsection{Base Model Agent}\label{app:sec:base-model-prompts}

The base model sub-graph (Fig.~\ref{fig:base-model-agent}) serves as a control condition, evaluating a vanilla LLM without multi-agent decomposition or RAG. When instantiated as First Speaker (Fig.~\ref{fig:parent-graph}), the agent generates an initial claim aligned with its assigned political stance. On subsequent turns, it produces protocol-compliant responses conditioned on its own and the opponent's initial claims, commitment sets, and dialogue history. Each invocation performs: (i) response generation; (ii) utterance classification; and (iii) commitment updates. Below we summarise the response-generation prompts.

\subsubsection{Task Framing}\label{app:subsubsec:base-model-task}

The agent is framed as a participant in a formal persuasion dialogue based on \citet{10.1017/S0269888906000865}. The dialogue is explicitly described as hypothetical and conducted under ethics approval to avoid refusal behaviour on political topics. The agent's objective is to persuade the opponent to adopt its standpoint through coherent argumentation.

\subsubsection{Formal Grounding}\label{app:subsubsec:base-model-definitions}

Key elements of the formal persuasion dialogue system are embedded directly in the system prompt:

\paragraph{Dialogue Goal and Structure.}
A persuasion dialogue resolves a conflict of viewpoints (positive, negative, or doubt) on a proposition. Resolution occurs when both participants share the same standpoint. The debate topic is explicitly specified.

\paragraph{Utterance Taxonomy.}
The six dialogue moves (App.~\ref{app:sec:persuasion-dialogues}) are defined with examples and a qualitative persuasiveness ranking to guide generations.

\paragraph{Win Condition.}
The agent wins if the opponent retracts their initial claim or concedes to the agent's initial claim.

\paragraph{Typical Response Rules.}
A mapping from each move type to its permissible replies (Tab.~\ref{tab:typical-replies-for-persuasio}) constrains response generation.

\subsubsection{Dynamic State Components}\label{app:subsubsec:base-model-dynamic}

At generation time, the system prompt integrates:

\begin{itemize}
    \item \textbf{Political alignment}: a natural-language description derived from the debate topic and stance parameters.
    \item \textbf{Initial claims and commitments}: both participants' commitment sets.
    \item \textbf{Dialogue history}: the full prior exchange.
    \item \textbf{Generic rules}: shared behavioural and formatting constraints.
\end{itemize}

\subsubsection{System Instructions}\label{app:subsubsec:base-model-instructions}

The agent is instructed to:

\begin{enumerate}
    \item Produce a persuasive response aligned with its initial claim, commitments, and political stance; if opening, generate a stance-consistent initial claim.
    \item Ensure the initial claim disagrees with the opponent's position if speaking second.
    \item Classify each opponent sentence by dialogue move type.
    \item Generate replies guided by the typical response rules.
\end{enumerate}

\subsubsection{User Message}\label{app:subsubsec:base-model-user-msg}

At each turn, the user message provides the opponent's most recent utterance (one to five sentences) and instructs the agent to respond persuasively in accordance with the system prompt and assigned political position.

\subsection{MAS Sub-Graph}\label{app:subsec:prompts:mas}

The MAS (Fig.~\ref{fig:mas}) decomposes persuasive response generation into a pipeline of specialised sub-agents, each implementing a distinct argumentative function. We hypothesised that separating generation and evaluation improves persuasiveness relative to the single-pass base model baseline. The pipeline consists of: (1) utterance classification (App.~\ref{app:subsec:prompts:utterance-classification}); (2) a protocol reasoner; (3) dialogue-move-specific generation agents; (4) a most persuasive choice agent; (5) a disambiguation agent; and (6) a commitment update agent (App.~\ref{app:subsec:prompts:commitment-updates}). Below we detail components not previously described.

\subsubsection{Dialogue Protocol Reasoner}\label{app:subsubsec:prompts:protocol-reasoner}

The protocol reasoner deterministically maps each opponent sentence to its admissible reply types using the formal response rules of \citet{10.1017/S0269888906000865} (Tab.~\ref{tab:typical-replies-for-persuasio}). Operating over utterance labels from the classification step, it returns the set of typical responses for each sentence. Each admissible type invokes its corresponding generation agent.

\subsubsection{Generation Agents}\label{app:subsubsec:prompts:generation-agents}

Generation agents are modular components, each responsible for a specific dialogue move type. For every admissible reply, the corresponding agent produces three candidate sentences per opponent sentence, creating a diverse candidate pool. System prompts share a common structure: political alignment (and vector-based persona if RAG-enabled), initial claims and commitments of both speakers, dialogue history, and formatting constraints.

\paragraph{Initial Claim Agent.}
Generates three one-sentence opening claims aligned with the assigned political stance. In the RAG setting, the vector-based persona string (App.~\ref{app:subsec:prompts:persona-gen}) conditions the initial claim generation on ideological orientation.

\paragraph{Claim Agent.}
Produces three stand-alone claims that answer an opponent's question or defend a prior commitment, ensuring coherence with dialogue history and commitment state.

\paragraph{Claim Negation Agent.}
Generates three direct contradictions to a specified opponent sentence, emphasising clarity and persuasive force.

\paragraph{Why Claim Agent.}
Produces three single-sentence questions beginning with ``why'' to request justification, thereby shifting the burden of proof to the opponent.

\paragraph{Question Claim Agent.}
Generates three journalistic questions (beginning with ``Who'', ``What'', ``When'', ``Where'', or ``How'') to probe weaknesses or elicit clarification.

\paragraph{Since Claim Agent.}
Produces three supporting premises that justify a prior claim, typically in response to a \textit{why} challenge, while maintaining commitment consistency.

\paragraph{Concede Claim Agent.}
Generates three concise concession statements (e.g., containing ``I accept'' or ``I admit'') explicitly acknowledging the opponent's proposition.

\paragraph{Retract Claim Agent.}
Produces three retraction statements withdrawing or clarifying prior commitments. This move does not use RAG, as the retrieval corpus of \cite{robinson2026validatingpoliticalpositionpredictions} does not contain retraction examples.

\subsubsection{Most Persuasive Choice Agent}\label{app:subsubsec:prompts:persuasiveness-choice}

After candidate generation, the most persuasive choice agent selects the final response set. Two prompt variants are used.

\paragraph{Initial Turn.}
Selects the single most persuasive opening claim based on tone, clarity, emotional impact, word choice, and call to action. Political alignment (and persona, if applicable) is embedded in the prompt.

\paragraph{Subsequent Turns.}
Selects one to five sentences from all generated candidates. The prompt includes: (i) a persuasiveness ranking of move types (\textit{concede} $=$ \textit{retract} $<$ \textit{why} $<$ \textit{question} $<$ \textit{since} $<$ \textit{claim}); (ii) political alignment (and persona if RAG-enabled); (iii) current claims and commitments; (iv) dialogue history; and (v) sentences already selected in the current turn. Criteria emphasise coherence with prior commitments and discourage repetition. Selected sentences must be preserved verbatim and form a jointly persuasive, stance-consistent argument.

\subsubsection{Disambiguation Agent}\label{app:subsubsec:prompts:disambiguation}

To remove redundancy among selected candidates, a three-stage disambiguation process is applied.

\paragraph{Semantic Grouping.}
An LLM groups sentences with equivalent or highly similar meanings into clusters, excluding singleton groups.

\paragraph{Representative Selection.}
For each cluster, the most persuasive sentence is selected and preserved verbatim via an LLM call.

\paragraph{Maximum Length Enforcement.}
If the remaining set exceeds a configurable maximum, a final selection step chooses exactly $k$ sentences using the same persuasiveness criteria (tone, clarity, emotional impact, word choice, call to action).

\subsection{Vector-Based Persona Generation}\label{app:subsec:prompts:persona-gen}

We extended the MAS with a RAG component (Fig.~\ref{fig:mas-rag}) built on the KB of \cite{robinson2026validatingpoliticalpositionpredictions}, which contains locution--proposition pairs, argumentative relations (e.g., attack, support, temporal transitions), and LLM-predicted political position scores. Candidate nodes were retrieved via cosine similarity over SBERT embeddings (\texttt{all-MiniLM-L6-v2}; \citealt{reimers-2019-sentence-bert}) and filtered using political alignment constraints derived from Ensemble~3 metrics. Left-wing personas were restricted to mean scores between 20--40 and right-wing personas to 60--80 (both with $\pm 10$ standard deviation), with non-political probability $\leq 0.25$.

For each debate topic, the top-$5$ most similar, stance-aligned nodes were incorporated into a structured prompt to generate a persona schema encoding ideological stance, values, policy positions, and communication style. Retrieved examples served as alignment signals rather than verbatim content and were updated to a 2025 context.

\subsubsection{Vector-Based Retrieval}\label{app:subsubsec:prompts:vector-retrieval}

At initialisation, two retrieval queries were issued: (i) the debate topic (if the agent was First Speaker; Fig.~\ref{fig:parent-graph}) or (ii) the opponent's initial utterance (if Second Speaker). Queries were embedded with \texttt{all-MiniLM-L6-v2} and matched against pre-computed node embeddings. Nodes were filtered as follows:

\begin{itemize}
    \item \textbf{Utterance type}: only \textit{claim}-labelled nodes were retained.
    \item \textbf{Political alignment}: mean position within 20--40 (left) or 60--80 (right), standard deviation $\leq 10$, and non-political probability $\leq 0.25$.
\end{itemize}

Remaining nodes were ranked by cosine similarity and the top-$5$ returned. For each node, both locutionary (surface text) and propositional forms were provided to supply rhetorical and semantic guidance.

\subsubsection{Persona Generation Prompt}\label{app:subsubsec:prompts:persona-prompt}

The LLM was prompted to construct a UK political persona schema using a system and user message.

\paragraph{System Message.}
The system prompt specified: (i) the debate topic, (ii) the opponent's initial utterance (if applicable), and (iii) a temporal constraint (current year: 2025; retrieved examples dated 2020--2021). The model was required to:
(i) output strictly in the \texttt{PersonaSchema} format,
(ii) avoid dialogue or multiple perspectives,
(iii) use retrieved examples only as alignment guides (no verbatim copying),
(iv) update outdated references while preserving ideological consistency, and
(v) remain within the assigned stance range. Format instructions were appended to enforce structured output.

\paragraph{User Message.}
The user prompt supplied one out of the follow two dynamically retrieved example sets depending on whether the agent was the First Speaker:
\begin{enumerate}
    \item Topic-aligned examples (top-$5$ by similarity to the debate topic).
    \item Opponent-aligned examples (top-$5$ by similarity to the opponent's utterance).
\end{enumerate}
Each example included indexed locutionary and propositional forms.

\subsubsection{Persona Schema}\label{app:subsubsec:prompts:persona-schema}

The \texttt{PersonaSchema} required the following fields: 
\textbf{political stance}, 
\textbf{main topic}, 
\textbf{core values}, 
\textbf{political positions on main topic}, 
\textbf{key policy positions}, 
\textbf{ideological framework} (political values, role of government, economic philosophy, social issues), 
\textbf{information sources} (preferred media, trusted figures, information processing), and 
\textbf{communication style} (discourse style, argumentation approach, rhetoric patterns, emotional triggers). All fields were mandatory.

\subsubsection{Persona String Construction}\label{app:subsubsec:prompts:persona-string}

Validated schemas were rendered into a natural-language persona string and prepended to all downstream system prompts. The string comprised: (i) \textbf{Identity} (UK political emulation with explicit stance range), (ii) \textbf{Persona Schema} (structured rendering of all fields), (iii) \textbf{Emulation Instructions} (remain in character, use British English, maintain belief consistency), and (iv) \textbf{Response Guidelines} (first-person voice, natural openings, anecdotal grounding, emotional coherence). This persona conditioning was injected into all generation agents, ensuring consistent and politically grounded dialogue behaviour.



\section{Utterance Classification Evaluation}\label{app:utt-class-eval}

To select an LLM for utterance classification in the human sub-graph, we evaluated candidate models on their ability to assign dialogue moves to illocutionary types within the persuasion framework.

\subsection{Dataset}

We constructed a manually annotated test set from the KB of \citet{robinson2026validatingpoliticalpositionpredictions}. Each utterance was labelled as one of: \textit{claim}, \textit{why}, \textit{since}, \textit{question}, or \textit{concede} (Tab.~\ref{tab:typical-replies-for-persuasio}). As the KB did not contain \textit{retract} moves, these were sampled from completed model--model dialogue transcripts. 

To balance classes, we sampled up to 10 instances per type, yielding 60 utterances in total.

\subsection{Classification Procedure}

All candidate models were evaluated using the same system prompt described in Appendix~\ref{app:subsec:prompts:utterance-classification}. The user prompt embedded the previous two dialogue turns to preserve chronological structure, which is necessary to distinguish \textit{claim} from \textit{since}:

\begin{quote}
\textit{Classify the following sentence according to the defined utterance types. Previous dialogue turns: User: ``A''; Model: ``B''. Current sentence: User: ``C''.}
\end{quote}

Chronology is critical because \textit{since} moves function as premises supporting a prior claim. To construct \textit{since} examples, we sampled nodes linked via \texttt{TRANSITIONS\_TO} relations (A $\rightarrow$ B $\rightarrow$ C) where A and C were produced by the same speaker and C supported A.

\subsection{Models and Parameters}

We evaluated GPT-4o, Grok~3, and Mistral Medium (Tab.~\ref{tab:models}). All models were run with near-deterministic parameters (temperature=$0$, \texttt{top\_p}=$1$, seed=$123$). Each model classified every sample 20 times to account for residual output variability.

\subsection{Bootstrap Evaluation}

We estimated performance using bootstrap resampling:

\begin{enumerate}
    \item 10{,}000 bootstrap samples (with replacement) of size $n=60$ were drawn.
    \item For each sample, we computed confusion matrices, per-class precision/recall/F1, and overall precision, recall, and macro-F1.
    \item Mean estimates and 95\% confidence intervals were derived from the bootstrap distributions.
\end{enumerate}

\subsection{Label-Set Configurations}

Because \textit{why} and \textit{question} are semantically similar and less frequent in the KB, we evaluated four label configurations:

\begin{enumerate}
    \item \textbf{Full}: all six types.
    \item \textbf{No Why}: excluding \textit{why}.
    \item \textbf{No Question}: excluding \textit{question}.
    \item \textbf{No Why and No Question}: excluding both.
\end{enumerate}

This ablation isolates the contribution of confusable categories to overall performance.

\subsection{Results}

\subsubsection{Full Label Set}

\begin{figure}[h]
    \centering
    \includegraphics[width=1\linewidth]{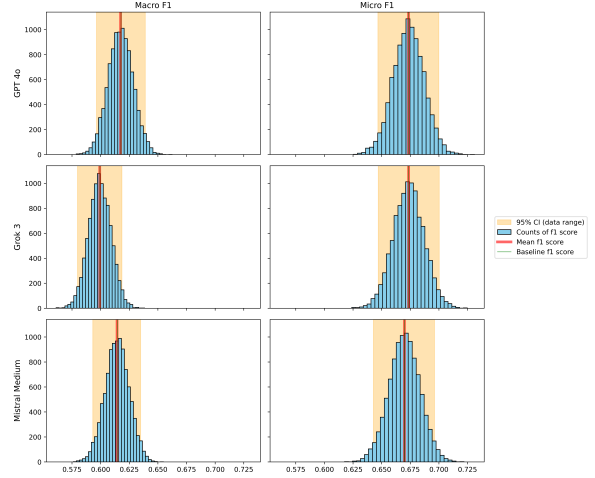}
    \caption{Bootstrapped macro and micro F1, and 95\% confidence interval ranges, for the three candidate models.}
    \label{app:fig:utt-class-bootstrapped-f1s-full-dataset}
\end{figure}

Under the full configuration (Fig.~\ref{app:fig:utt-class-bootstrapped-f1s-full-dataset}), macro-F1 scores clustered around 0.60--0.62 across models, with micro-F1 slightly higher (0.65--0.67). Bootstrap distributions were approximately normal, indicating stable performance under resampling. Confidence intervals (width $\approx$ 0.05--0.08) reflect variance due to the limited test set. GPT-4o achieved the highest macro-F1, with Grok~3 and Mistral Medium performing comparably.

\subsubsection{Label-Set Ablations}

\begin{figure}[h]
    \centering
    \includegraphics[width=1\linewidth]{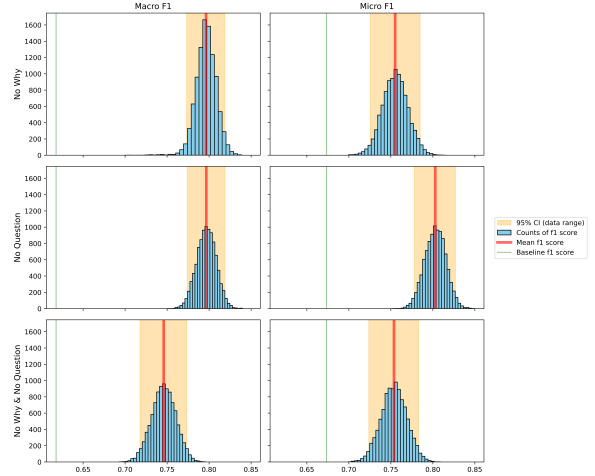}
    \caption{Label-set ablations for GPT 4o, the best performing candidate model for utterance classification. The green baseline F1 scores are computed on the full (non-ablated) label set for reference.}
    \label{app:fig:utt-class-bootstrapped-f1s-full-dataset-ablations}
\end{figure}

Figure~\ref{app:fig:utt-class-bootstrapped-f1s-full-dataset-ablations} shows ablations for GPT-4o.

\paragraph{No Why.}
Removing \textit{why} increased mean F1 to approximately 0.76--0.78 with narrower confidence intervals, indicating that \textit{why} was a major source of confusion, likely due to overlap with \textit{question}.

\paragraph{No Question.}
Excluding \textit{question} similarly improved F1 (0.72--0.76), suggesting symmetric confusion between these interrogative categories.

\paragraph{No Why and No Question.}
Removing both yielded stable performance around 0.74--0.76, confirming that classification difficulty was concentrated in these two types. The reduced label set (\textit{claim}, \textit{since}, \textit{concede}, \textit{retract}) was the most reliable.

\paragraph{Discussion.}
Across ablations, F1 scores exceeded those of the full configuration, confirming that performance degradation was driven by confusion between \textit{why} and \textit{question}, rather than bootstrap artefacts.

\paragraph{Model Selection.}
GPT-4o was selected as the utterance classifier for the human sub-graph (Fig.~\ref{fig:human-agent}) due to its marginally superior and more stable macro-F1. Although ablations improved reliability, the full label set was retained during experiments because \textit{why} and \textit{question} moves occurred in human dialogues. The ablation results informed interpretation, as misclassification between these categories was anticipated.

\section{Annotation Guidelines}\label{app:subsec:annotation-guidelines}

The following instructions were presented to Prolific annotators for the subjective persuasiveness evaluation task.

\subsection{Task Overview}\label{app:subsubsec:annotation-task-overview}

Annotators evaluated pairs of debate transcripts on the same topic. Each transcript featured Speaker A and Speaker B. Although the individuals differed across debates, Speaker A represented the same ideological position in both. Annotators were asked to compare the persuasiveness of Speaker A across the two debates, focusing exclusively on argumentative effectiveness.

They were informed that, aside from attention checks, there were no correct or incorrect answers and that honest, subjective judgements were required.

\subsection{Task Procedure}\label{app:subsubsec:annotation-task-steps}

For each instance, annotators were instructed to:

\begin{enumerate}
    \item Review the \textbf{debate topic}.
    \item Read \textbf{Debate 1}.
    \item Read \textbf{Debate 2}.
    \item Compare the two \textbf{Speaker A} participants.
    \item Select which Speaker A was more persuasive overall.
    \item Rate their \textbf{confidence} in this judgement.
    \item Complete any \textbf{attention checks}.
\end{enumerate}

\subsection{Important Instructions}\label{app:subsubsec:annotation-notes}

Annotators were reminded to:

\begin{itemize}
    \item Compare only the two Speaker A participants.
    \item Base decisions solely on transcript content.
    \item Complete two mandatory attention checks per session.
    \item Provide responses that exactly matched any instructions appearing in parentheses, e.g. ``(ATTENTION CHECK: \ldots)''.
\end{itemize}

Submissions failing attention checks were returned. All other responses were evaluated as subjective judgements.

\subsection{Annotation Questions}\label{app:subsubsec:annotation-questions}

Each instance consisted of two questions:

\paragraph{Q1: Persuasiveness.}
\textit{“Based only on the debate transcripts, which Speaker A presented the more persuasive arguments?”}
\begin{itemize}
    \item Speaker A from Debate 1
    \item Speaker A from Debate 2
    \item Both were equally persuasive
\end{itemize}

\paragraph{Q2: Confidence.}
\textit{“How confident are you in your choice?”}
\begin{itemize}
    \item Very unsure
    \item Somewhat unsure
    \item Neutral
    \item Somewhat confident
    \item Very confident
\end{itemize}

\end{document}